\documentclass[11pt]{article}
\usepackage[margin=1in]{geometry}
\usepackage[utf8]{inputenc}
\usepackage{url,comment,amsmath,booktabs,graphicx,listings}
\usepackage[ruled,linesnumbered]{algorithm2e}

\SetCommentStyOne{mycommfontone}
\SetCommentStyTwo{mycommfonttwo}
\usepackage[margin=1cm,font={small,sf,bf}]{caption}
\usepackage{subcaption}
\usepackage{siunitx}
\usepackage{hyperref}
\usepackage[table,xcdraw]{xcolor}
\usepackage{authblk}
\usepackage{amssymb}

\usepackage{color, colortbl}
\definecolor{Gray}{gray}{0.9}

\usepackage{etoolbox}
\makeatletter
\patchcmd{\@algocf@start}
  {-1.5em}
  {0pt}
  {}{}
\makeatother

\usepackage[backend=biber,style=nature,]{biblatex}
\newcommand\myurl[1]{\small{\url{#1}}}

\newif\iffinal
\finaltrue

\iffinal
    \newcommand\ben[1]{}
    \newcommand\ian[1]{}
    \newcommand\status[1]{}
    \newcommand\note[1]{}
    \newcommand\hong[1]{}
    \newcommand\kyle[1]{}

\else
    \newcommand\ben[1]{{\color{blue}[Ben: #1]}}
    \newcommand\ian[1]{{\color{red}[Ian: #1]}}
    \newcommand\status[1]{{\color{purple}[Status \today{}: #1]}}
    \newcommand\note[1]{{\color{purple}[Note: #1]}}
    \newcommand\hong[1]{{\color{cyan}[Hong: #1]}}
    \newcommand\kyle[1]{{\color{green}[Kyle: #1]}}
\fi

\title{AI- and HPC-enabled Lead Generation for SARS-CoV-2:\\Models and Processes to Extract Druglike Molecules Contained in Natural Language Text}
\author[1]{Zhi Hong}
\author[1]{J. Gregory Pauloski}
\author[2]{Logan Ward}
\author[1,2]{Kyle Chard}
\author[3,2]{Ben Blaiszik}
\author[1,2]{Ian Foster}

\affil[1]{Department of Computer Science, University of Chicago}
\affil[2]{Data Science and Learning Division, Argonne National Laboratory}
\affil[3]{Globus, University of Chicago, Chicago, IL, USA}

\begin{document}

\maketitle

\abstract{
\noindent{}Researchers worldwide are seeking to repurpose existing drugs or discover new drugs to counter the disease caused by severe acute respiratory syndrome coronavirus 2 (SARS-CoV-2). A promising source of candidates for such studies is molecules that have been reported in the scientific literature to be drug-like in the context of coronavirus research. 
We report here on a project that leverages both human and artificial intelligence to detect references to drug-like molecules in free text.
We engage non-expert humans to create a corpus of labeled text, use this labeled corpus to train a named entity recognition model, and employ the trained model to extract \num{10912} drug-like molecules from the COVID-19 Open Research Dataset Challenge (CORD-19) corpus of \num{198875} papers. Performance analyses show that our automated extraction model can achieve performance on par with that of non-expert humans.
}

\section{Introduction}
The Coronavirus Disease (COVID-19) pandemic, caused by transmissible infection of the severe acute respiratory syndrome coronavirus 2 (SARS-CoV-2), has resulted in tens of millions of diagnosed cases and over \num{1450000} deaths worldwide~\cite{dong2020interactive}; straining healthcare systems, and disrupting key aspects of society and the wider economy. 
It is thus important to identify effective treatments rapidly via discovery of new drugs and repurposing of existing drugs. Here, we leverage advances in natural language processing to enable automatic identification of drug candidates being studied in the scientific literature.

The magnitude of the pandemic has resulted in an enormous number of academic publications related to COVID-19 research since early 2020. Many of these articles are collated in the COVID-19 Open Research Dataset Challenge (CORD-19) collection \cite{cord19,wang2020cord}.
With \num{198875} articles at the time of writing, that collection is far too large for humans to read.
Thus, tools are needed to automate the process of extracting relevant data, such as drug names, testing protocols, and protein targets. 
Such tools can save domain experts significant time and effort.

Towards this goal, we describe here how we have tackled two important problems: creating labelled training data via judicious use of scarce human expertise, and applying a named entity recognition (NER) model to automatically identify drug-like molecules in text. 
In the absence of rich labeled data for the growing COVID literature, we employ an iterative model-in-the-loop collection process 
inspired by our previous work~\cite{tchoua2019active,tchoua2019creating}. 
We first assemble a small bootstrap set of human-verified examples to train a model for identifying similar examples.
We then iteratively apply the model, use human reviewers to 
verify the predictions for which the model is least confident, and 
retrain the model until the improvement in performance is less than
a threshold.
(The human reviewers were administrative staff without scientific backgrounds,
with time available for this task due to the pandemic.)

Having collected adequate training data via this model-guided human annotation process,
we then use the resulting labeled data to re-train a NER model originally developed to identify polymer names in materials science publications \cite{hong2020sciner} and apply this trained model to CORD-19. We show that the labeled data produced by our approach are of sufficiently high quality than when used to train NER models, which achieves a best F-1 score of 80.5\%---roughly equivalent to that achieved by non-expert humans.

The labeled data, model, and model results are all available online, as described in Section~\ref{sec:data_availability}.

\section{Problem Definition}
We aim to develop and apply new computational methods to mine the scientific literature to identify small molecules
that have been investigated or found useful as antiviral therapeutics. 
For example, processing the following sentence should allow us to determine that the drug \emph{sofosbuvir}
has been found effective against the Zika virus:
``Sofosbuvir, an FDA-approved nucleotide polymerase inhibitor, can efficiently inhibit replication and infection of several ZIKV strains, including African and American isolates.'' \cite{bullard2017fda}.

This problem of identifying drug-like molecules in text can be divided into two linked problems: 
1) identifying references to small therapeutic molecules (``drugs'') and 
2) determining what the text says about those molecules.
In this work, we consider potential solutions to the first problem.

A simple way to identify entities in text that belong to a specialized class
(e.g., drug-like molecules)
is to refer to a curated list of valid names, if such is available.
In the case of drugs, we might think to use
DrugBank \cite{wishart2018drugbank} or the FDA Drug Database \cite{drugsatfda},
both of which in fact list \emph{sofosbuvir}.
However, such databases are not in themselves an adequate solution to our problem,
for at least two reasons.
First, they are rarely complete. 
The tens of thousands of entity names in DrugBank and the FDA Drug Database together are just a tiny fraction of the
billions of molecules that could potentially be used as drugs. 
Second, such databases may be overly general: 
DrugBank, for example, includes the terms ``rabbit'' and ``calcium,'' 
neither of which have value as antiviral therapeutics.
In general, the use of any such list to identify entities 
will lead to both false negatives and false positives.
We need instead to employ the approach that a human reader might follow in this situation, namely to scan text for words that appear in contexts in which a drug name is likely to appear.
In the following, we explain how we use natural language processing (NLP) techniques for this purpose.

\section{Automated Drug Entity Extraction from Literature}

Finding strings in text that refer to drug-like molecules is an example of named-entity recognition (NER) \cite{nadeau2007survey},
an important NLP task.  
Both grammatical and statistical (e.g., neural network-based) methods have been applied to NER;
the former can be more accurate, 
but require much effort from trained linguists
to develop.
Statistical methods use supervised training on labeled examples to 
learn the contexts in which entities of interest (e.g., drug-like molecules) are likely to occur, and then classify previously unseen words as such entities if they appear in similar contexts. 
For instance, a training set may contain the sentence ``Ribavirin was administered once daily by the i.p.\ route'' \cite{oestereich2014evaluation}, with \emph{ribavirin} labelled as \textit{Drug}. With sufficient training data, the model may learn to assign the label \textit{Drug} to \emph{arbidol} in the sentence ``Arbidol was administered once daily per os using a stomach probe'' \cite{oestereich2014evaluation}. This learning approach can lead to general models capable of finding previously unseen candidate molecules in natural language text.

The development of effective statistical NER models is complicated by the many contexts in which names can occur.
For example, while the contexts just given for \emph{ribavirin} and \emph{arbidol} are similar,
both are quite different from that quoted for \emph{sofosbuvir} earlier.
Furthermore, authors may use different 
wordings and sentence structures: e.g., ``given by i.p.\ injection once daily''
rather than ``administered once daily by the i.p.\ route.''
Thus, statistical NER methods need to do more than learn template word sequences:
they need to learn more abstract representations of the context(s) in which words appear. Modern NLP and NER systems do just that \cite{chiu2016named}.

\subsection{SpaCy and Keras-LSTM Models}\label{sec:models}

We consider two NER models in this paper, 
SpaCy and a Keras long-short term memory (LSTM) model.
Both models are publicly available on DLHub~\cite{li147dlhub} and GitHub,
as described in Section~\ref{sec:data_availability}.

SpaCy is an open source NLP library that provides a pre-trained entity recognizer that can recognize 18 types of entities, including PERSON, ORGANIZATION, LOCATION, and PRODUCT. Its model calculates a probability distribution of a word over the entity types, and outputs the type with the highest probability as the predicted type for that word. 
When pre-trained on the OntoNotes~5 dataset of over 1.5 million labeled words \cite{weischedel2013ontonotes}, the SpaCy entity recognizer can identify supported entities with 85.85\% accuracy. However, it does not include drug names as a supported entity class, and thus we would need to retrain the SpaCy model on a drug-specific training corpus.
Unfortunately, there is no publicly available corpus of labeled text for drug-like molecules in context.
Thus, we need to use other methods to retrain this model (or other NER models), as we describe in Section~4.

While SpaCy is easy to use, it lacks flexibility: its end-to-end encapsulation does not expose many tunable parameters. Thus we also explore the use of a Keras-LSTM model that we developed in previous work for identification of polymers in materials science literature \cite{hong2020sciner}. 
This model is based on the Bidirectional LSTM network with a conditional random field (CRF) layer added on top. It takes training data labeled according to the ``IOB'' schema. The first word in an entity is given the label ``B'' (Beginning), the following words in the same entity are labeled ``I'' (Inside), and non-entity words are labeled ``O'' (outside). During prediction, the Bi-LSTM network tries to assign one of ``IOB'' to each word in the input sentence, but it has no awareness of the validity of the label sequence. The CRF layer is used on top of Bi-LSTM to lower the probability of invalid label sequences (e.g., ``OIO''). 

We compare the performance of SpaCy and Keras-LSTM models under various conditions in Section~\ref{sec:tradeoff}.

\subsection{Model-in-the-Loop Annotation Workflow}
\label{sec:data_collection}

\begin{figure}[bht]
    \centering
    \includegraphics[width=0.89\linewidth,trim=5mm 49mm 4mm 32mm,clip]{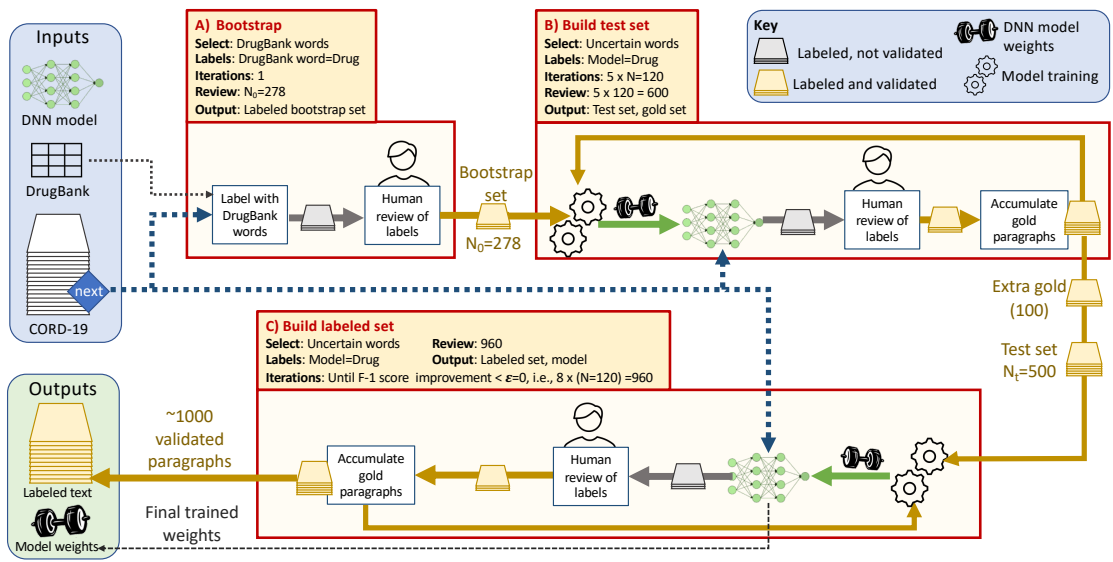}
    \caption{Overview of the training data collection workflow, showing the three  phases described in the text and with the parameter values used in the study reported here.
    Each phase pulls paragraphs from the CORD-19 dataset (blue dashed line) according to the \textbf{Select} criterion listed in its shaded box.
    Phases B and C repeatedly update the weights for the NER model (green arrows) that they use to identify and label uncertain paragraphs; human review (yellow and gold arrows) corrects those silver paragraphs to yield gold paragraphs. Total human review work is $\sim$278+600+960=1838 paragraphs. \kyle{In the boxes does review mean how many things people review? Or is it the number of positive reviews? Seems more interesting to show there how many paragraphs the human reviewers reviewed, but at least for A and B it seems to indicate the number of validated outputs}}
    \label{fig:labeling_workflow}
\end{figure}

We address the lack of labeled training data by using Algorithm~\ref{alg:workflow} (and see Figure~\ref{fig:labeling_workflow})
to assemble a set of human- and machine-labeled data from CORD-19 \cite{wang2020cord}.
In describing this process, we refer to paragraphs labeled automatically via a heuristic or model as \emph{silver} and to silver paragraphs for which labels have been corrected by human reviewers as \emph{gold}. 
We use the Prodigy machine learning annotation tool to manage the review process:
reviewers are presented with a silver paragraph, with putative drug entities highlighted; they click on false negative and false positive words to add or remove the highlights and thus produce a gold paragraph. Prodigy saves the corrected labels in standard NER training data format.

Our algorithm involves three main phases, as follows.
In the first \textbf{bootstrap} phase, 
we
assemble an initial test set of gold paragraphs for use in
subsequent data acquisition.
We create a first set of silver paragraphs by using a simple heuristic:
we select $N_0$ paragraphs from CORD-19 that contain one or more words 
in DrugBank with an Anatomical Therapeutic Chemical Classification System (ATC) code,
label those words as drugs, and ask human reviewers to correct both false positives and false negatives in our silver paragraphs, creating gold paragraphs.
In the subsequent \textbf{build test set} phase,
we repeatedly 
use all gold paragraphs obtained so far to train an NER model;
use that model to identify and label additional silver paragraphs,
and engage human reviewers to correct false positives and false negatives, creating additional gold paragraphs.
We repeat this process until we have $N_t$ initial gold paragraphs.

In the third \textbf{build labeled set} phase,
we repeatedly use an NER model trained on all human-validated labels obtained to date,
with the $N_t$ gold paragraphs from the bootstrap phase used as a test set,
to identify and label promising paragraphs in CORD-19 for additional human review.
To maximize the utility of this human effort, 
we present the reviewers only with paragraphs that contain one or more \emph{uncertain words}, i.e., words that the NER model 
identifies as drug/non-drug with a confidence in the range [\texttt{min}, \texttt{max}]).
We continue this process of model retraining, paragraph selection and labeling,
and human review until the F-1 score improves by less than $\epsilon$.

The behavior of this algorithm is influenced by six parameters:
$N_0$, $N$, $N_t$, $\epsilon$, \texttt{min}, and \texttt{max}. 
$N_0$ and $N$ are the number of paragraphs that are assigned to human reviewers in the first and subsequent steps, respectively. 
$N_t$ is the number of examples in the test set.
$\epsilon$ is a threshold that determines when to stop collecting data. 
The \texttt{min} and \texttt{max} determine the confidence range 
from which words are selected for human review.
In the experimental studies described below, 
we used $N_0$=278, $N$=120,
$N_t$=500, $\epsilon$=0,
\texttt{min}=0.45, and \texttt{max}=0.55.

The NER model used in the model-in-the-loop
annotation workflow to score words might also be viewed as a parameter.
In the work reported here, we use SpaCy exclusively for that purpose, 
as it integrates natively with the Prodigy annotation tool and trains more rapidly.
However, as we show below, the Keras-LSTM model is ultimately somewhat more accurate
when trained on all of the labeled data generated, 
and thus is preferred when processing the entire CORD-19 dataset: 
see Section~\ref{sec:spacy_keras_performance} and Section~\ref{sec:apply}.

\vspace{1ex}

{\centering
\begin{minipage}{\linewidth}
    \begin{algorithm}[H]
        \SetAlgoLined
        ${\cal{C}}$ := \textbf{randomized\_paragraphs}(CORD) \tcp*{Prepare CORD-19 paragraphs}\label{line:1}
        \tcc{\textbf{A) Bootstrap:~Identify first silver paragraphs}}
        ${\cal{D}} := \{s : s \in$ DrugBank \& ATC($s$)\} \tcp*{Extract names from DrugBank}
        ${\cal{P}}_0 := \{ p : p =$ \textbf{next}$({\cal{C}})\ \&\ p \cap {\cal{D}} \neq \varnothing\}$ \& $|{\cal{P}}_0|=N_0$ \tcp*{Assemble first silver set}
        ${\cal{B}}$ = \textbf{verify}(${\cal{P}}_0$) \tcp*{Verify labels; initialize bootstrap set}
        \tcc{\textbf{B) Build test set:~Expand on bootstrap set from (A)}}
        \While{${|{\cal{B}}|<N_t}$}{
            ${\cal{M}} := \textrm{NER}(\textrm{train}=0.6 {\cal{B}}, \textrm{test}=0.4 {\cal{B}})$ \tcp*{Train: 60-40 train-test split}
            ${\cal{P}} := \{ p : p =$ \textbf{next}$({\cal{C}})\ \&\ \exists w \in p : {\cal{M}}(w) \in [\texttt{min}, \texttt{max}]) \}\ \&\ |{\cal{P}}|=N$ \tcp*{Next silver}
            ${\cal{V}}$ := \textbf{verify}(${\cal{P}}$) \tcp*{Engage crowd to verify labels}
            ${\cal{B}} := {\cal{B}} \cup {\cal{V}}$ \tcp*{Add corrected paragraphs to bootstrap set}
        }
        ${{\cal{T}} = {\cal{B}}[:N_t]}$ \tcp*{Initialize test set to first $N_t$ examples}
        ${{\cal{G}} = {\cal{B}}[N_t:]}$ \tcp*{Initialize gold set to any remaining examples}
        \tcc{C) Build labeled set:~Use test set from (B) to evolve model}
         \While{F-1 score improvement $> \epsilon$}{
            ${\cal{M}} := \textrm{NER}(\textrm{train}={\cal{G}}, \textrm{test}={\cal{T}})$ \tcp*{Train on ${\cal{G}}$, test on ${\cal{T}}$}
            ${\cal{P}} := \{ p : p =$ \textbf{next}$({\cal{C}})\ \&\ \exists w \in p : {\cal{M}}(w) \in [\texttt{min}, \texttt{max}]) \}\ \&\ |{\cal{P}}|=N$ \tcp*{Next silver}
            ${\cal{V}}$ := \textbf{verify}(${\cal{P}}$) \tcp*{Engage crowd to verify labels}
            ${\cal{G}} := {\cal{G}} \cup {\cal{V}}$ \tcp*{Add corrected paragraphs to gold set}
        }
        \caption{Model-in-the-loop Annotation Workflow}
        \label{alg:workflow}
    \end{algorithm}
\end{minipage}
\par
}

\vspace{1ex}

This semi-automated method saves time and effort for human reviewers because they are only asked to verify labels that have already been identified by our model to be uncertain, and thus worth processing.
Furthermore, as we show below, we find that we do not need to engage biomedical professionals to 
label drugs in text: untrained people, armed with contextual information (and online search engines), can spot drug names in text with accuracy comparable to that of experts. 

We provide further details on the three phases of the algorithm in the following, with numbers in the list referring to line numbers in Algorithm~\ref{alg:workflow}. 
\kyle{The following might be clearer if you include numbers for everything.}

\vspace{1ex}

\noindent
\textbf{A) Bootstrap}
\begin{enumerate}
    \item[1]
    We start with the 2020-03-20 release version of the CORD-19 corpus, which contains \num{44220} papers~\cite{wang2020cord}.
    We create ${\cal{C}}$, a random permutation of its paragraphs from which we will repeatedly fetch paragraphs via \texttt{next}(${\cal{C}}$).  
    \item[2]
    We bootstrap the labeling process by identifying as ${\cal{D}}$ the 2675 items in the DrugBank ontology with a Anatomical Therapeutic Chemical Classification System (ATC) code attached (eliminating many, but not all, druglike molecule entities).
    \item[3]
    We create an initial set of silver paragraphs, ${\cal{P}}_0$, by selecting $N_0$ paragraphs from ${\cal{C}}$ that include a word from ${\cal{D}}$.
    \item[4]
    We engage human reviewers to remove false positives and label false negatives in ${\cal{P}}_0$, yielding an initial set of gold paragraphs, ${\cal{B}}$.
\end{enumerate}
\noindent
\textbf{B) Build test set}
\begin{enumerate}
    \item[5]
    We expand the test set that we will use to evaluate the model created in the next phase, until we have $N_t$ validated examples.
    \item[6] \label{step:train}
    We train the NER model on 60\% of the data collected to date and evaluate it on the remaining 40\%,
    to create a new trained model, ${\cal{M}}$, with improved knowledge of the types of entities that we seek.
    \item[7] \label{step:new_silver}
    We use the probabilities over entities returned by the model to select, as our $N$ new 
    silver paragraphs, ${\cal{P}}$, 
    paragraphs that contain at least one uncertain word (see above).
    \item[8] \label{step:silver_to_gold}
    We engage human reviewers to convert these new silver paragraphs, ${\cal{P}}$, to gold, ${\cal{V}}$.
    \item[9] \label{step:add_new_to_gold} 
    We add the new gold paragraphs, ${\cal{V}}$, to the bootstrap set ${\cal{B}}$.
    \item[11--12] 
    Having assembled at least $N_t$ validated examples, 
    we select the first $N_t$ as the test set, $\cal{T}$, and
    use any remaining examples to initialize the new gold set, ${\cal{G}}$.
    \end{enumerate}
\noindent
\textbf{C) Build labeled set} 
\begin{enumerate}
    \item[13] 
    We assemble a training set ${\cal{G}}$, using the test set ${\cal{T}}$ assembled in the previous phases for testing.
    This process continues until the F-1 score stops improving (see Section~\ref{sec:tradeoff}). 
    \item[14--17] 
    Same as Steps~6--9, except that 
    we train on 
    ${\cal{G}}$ and test on ${\cal{T}}$,
    and add new gold paragraphs to ${\cal{G}}$ instead of ${\cal{T}}$.
    \end{enumerate}

\section{Data-Performance Tradeoffs in NER Models}
\label{sec:tradeoff}

As noted in Section~\ref{sec:data_collection}, 
our model-in-the-loop annotation workflow requires repeated retraining of a SpaCy model.
Thus we conducted experiments to understand how SpaCy prediction performance 
is influenced by model size, quantity of training data, and amount of training performed.

As the training data produced by the model-in-the-loop evaluation workflow
are to be used to train an NER model that we will apply to the entire CORD-19 dataset, we also evaluate the Keras-LSTM model from the perspectives of big data accuracy and training time.

\subsection{Model Size}
We first need to decide which SpaCy model to use for model-in-the-loop annotation. Model size is a primary factor that affects training time and prediction performance. In general, larger models tend to perform better, but require both more data and more time to train effectively.
As our model-in-the-loop annotation strategy requires frequent model retraining,
and furthermore will (initially at least) have little data,
we hypothesize that a smaller model may be adequate for our purposes.

\begin{figure}[bht]
    \centering
    \includegraphics[width=0.6\linewidth]{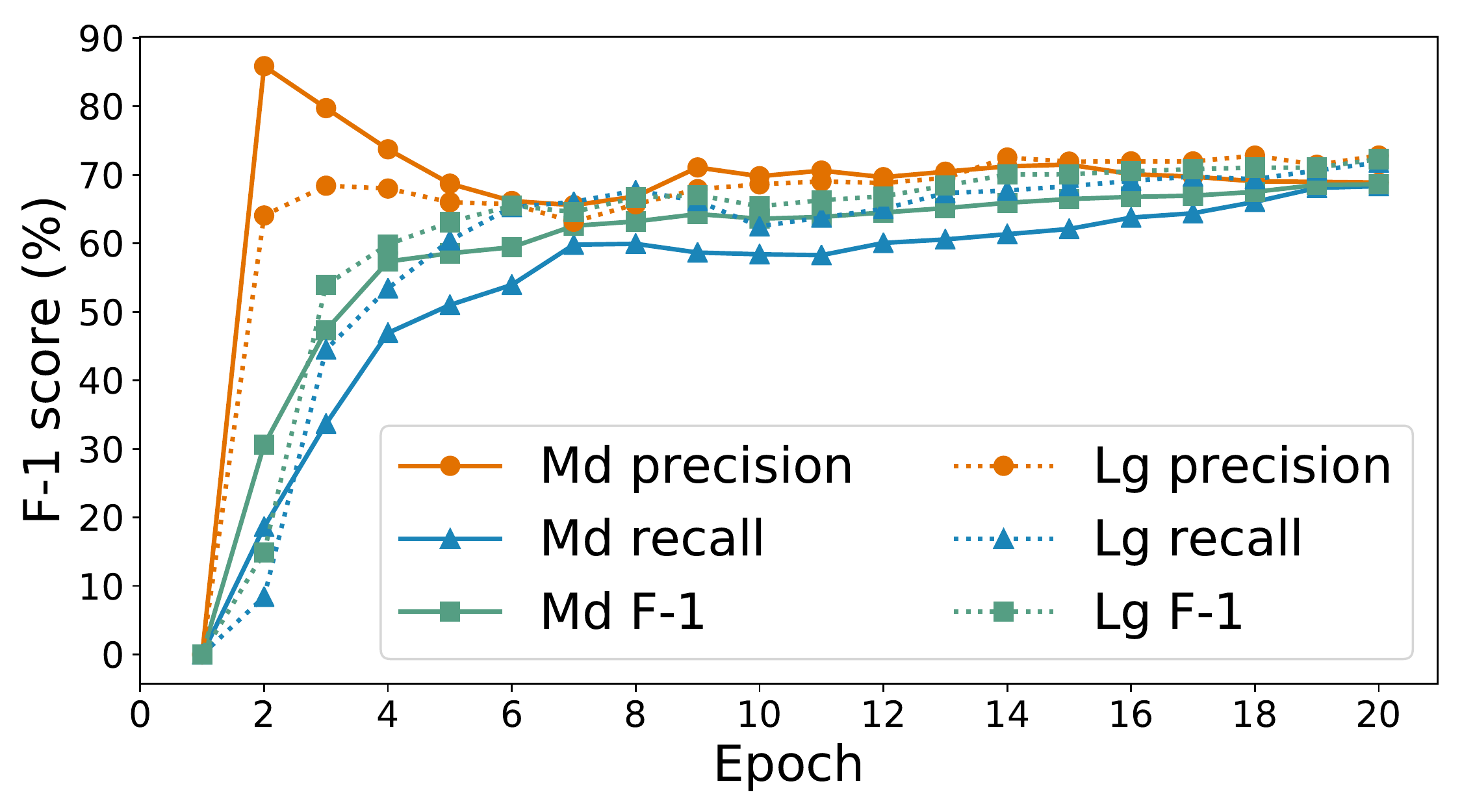}
    \caption{Precision, recall, and F-1 of medium and large SpaCy models trained on 278 examples.}
    \label{fig:lg_md}
\end{figure}

To explore this hypothesis, we study the performance achieved by the SpaCy medium and large models on our initial training set of 278 labeled paragraphs.
We show in Figure~\ref{fig:lg_md} the performance achieved by the two models as a function of number of training epochs.
Focusing on the harmonic mean of precision and recall, the F-1 score (a good measure 
a model's ability to recognize both true positives and true negatives), 
we see that the two models achieve similar prediction performance, with the largest difference in F-1 score being around 2\%. 
As the large model takes over eight times longer to train per epoch, we select the medium model for model-in-the-loop data collection.

\begin{figure}[bht]
    \centering
    \includegraphics[width=0.6\linewidth]{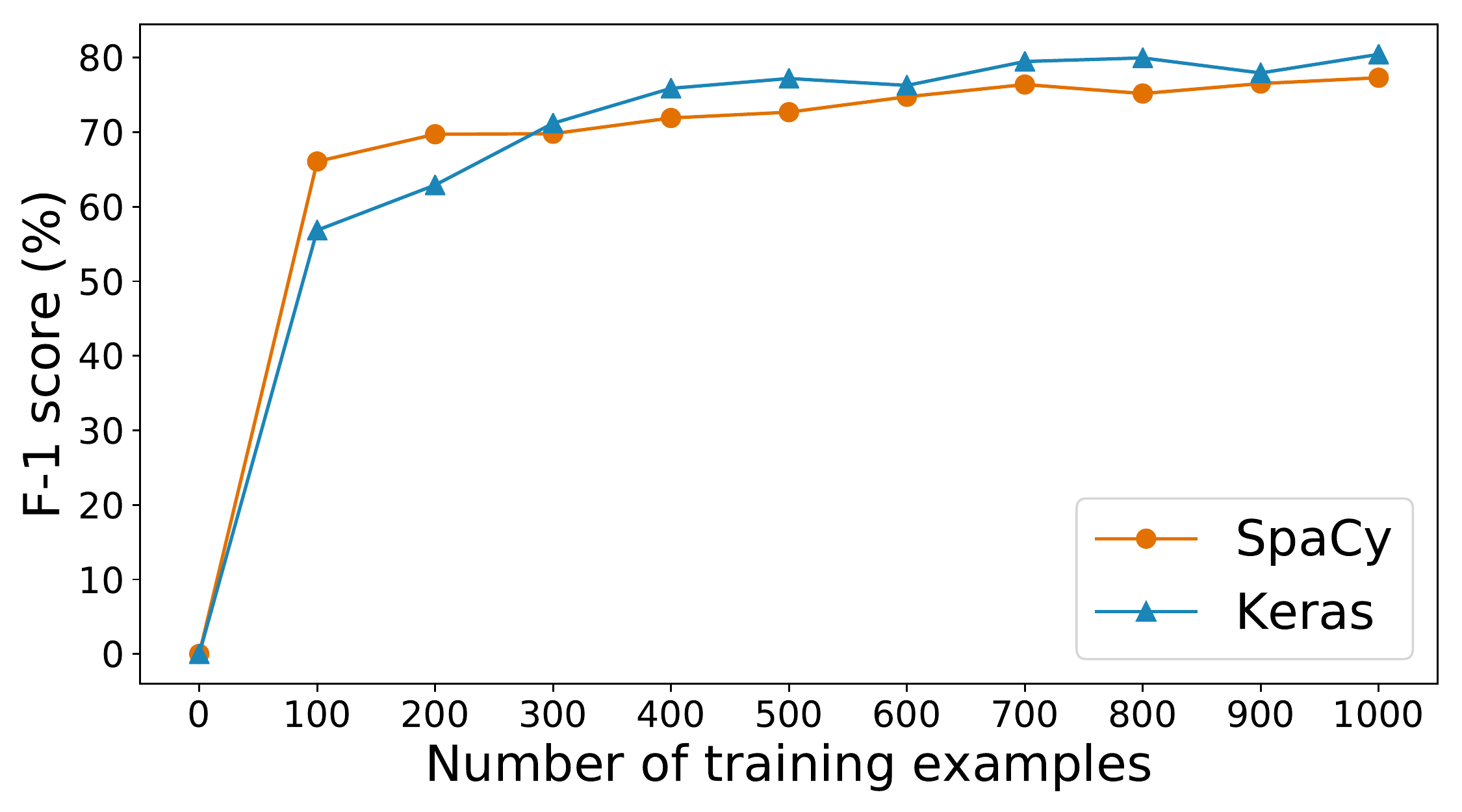}
    \caption{Training curve with different number of examples collected.}
    \label{fig:num_of_ex}
\end{figure} 

\subsection{Amount of Training Data}
\label{sec:amount_of_data}
As data labeling is expensive in both human time and model training time,
it is valuable to explore the tradeoff between time spent collecting data and prediction performance.
To this end, we manually labeled a set of 500 paragraphs selected at random from CORD-19 \cite{wang2020cord} as a test set. 
Then, we used that test set to evaluate the results of training the SpaCy and Keras-LSTM models of Section~\ref{sec:models} on increasing numbers of the paragraphs produced by our human-in-the-loop annotation process.
Figure~\ref{fig:num_of_ex} shows their F-1 score curves as we scale from 0 to 1000 training samples.
With only 100 training examples, SpaCy and Keras-LSTM achieve F-1 scores of 57\% and 66\%, respectively. SpaCy performs better than Keras-LSTM with fewer training examples (i.e., less than 300), after which Keras-LSTM overtakes it and maintains a steady 2--3\% advantage as the number of examples increases.
This result motivates our choice of Keras-LSTM for the CORD-19 studies in Section~\ref{sec:apply}.

We stopped collecting training data after \num{1000} examples. We see in Figure~\ref{fig:num_of_ex} that the performance of the SpaCy and Keras-LSTM models is essentially the same with 1000 training examples as with 700 examples, with the F-1 score even declining when the number of available examples increases to 800 or 900. At \num{1000} examples the F-1 score is greatest for both models. We conclude that the 1000 training examples, along with the other 500 withheld as the test set, are best-suited to train our models.

\subsection{Training Epochs}

Prediction performance is also influenced by the number of epochs spent in training. The cost of training is particularly important in a model-in-the-loop setup, as human reviewers cannot work while an model is offline for training.

\begin{figure}[bht]
    \centering
    \includegraphics[width=0.6\linewidth]{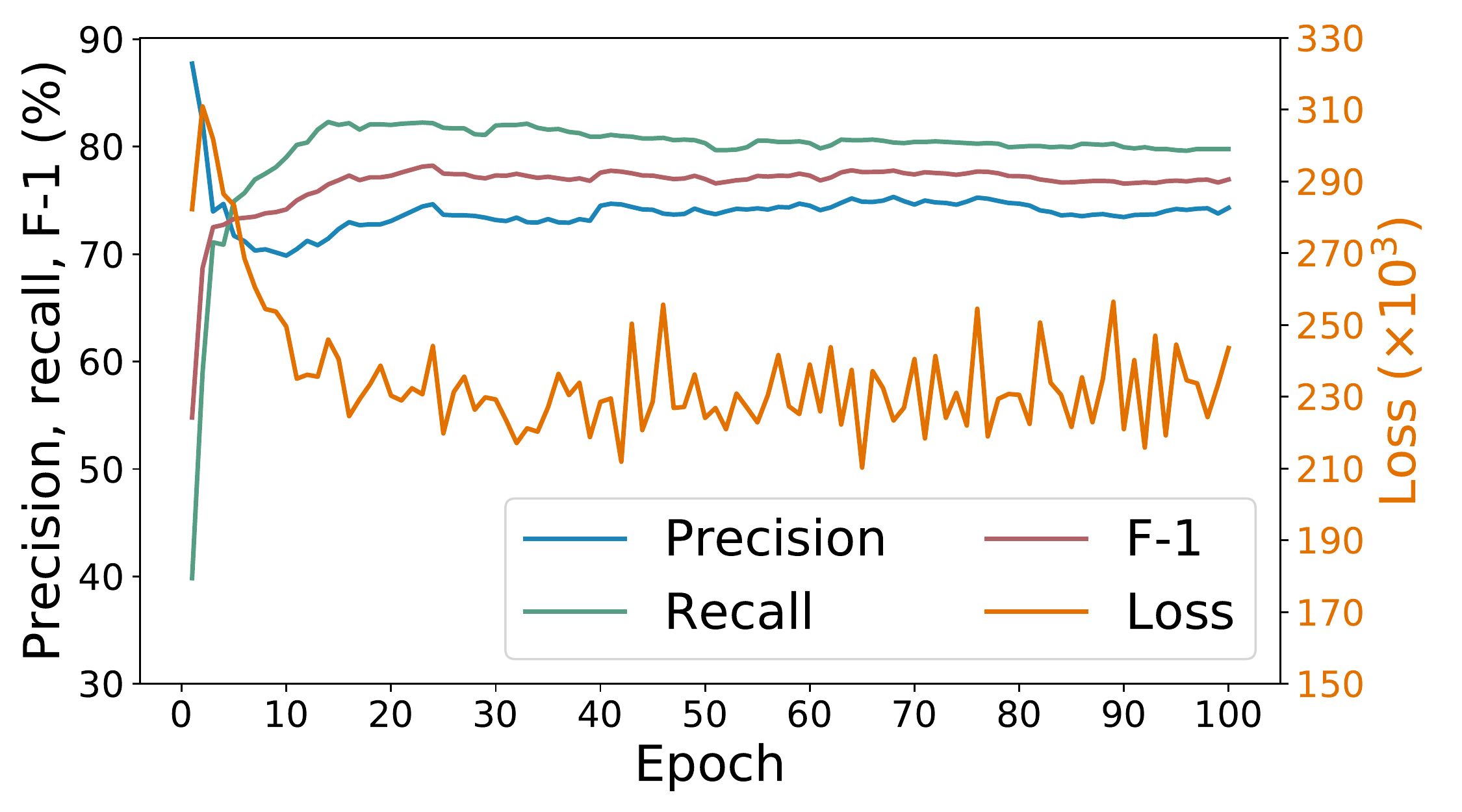}
    \caption{Loss, precision, recall, and F-1 of SpaCy model during training for 100 epochs on 278 paragraphs.}
    \label{fig:spacy_epoch}
\end{figure}

Figure~\ref{fig:spacy_epoch} shows the progression of the loss, precision, recall, and F-1 values of the SpaCy model during 100 epochs of training with the initial 278 examples. We can see that the best F-1 score is achieved within 10 to 20 epochs. Increasing the number of epochs does not result in any further improvement.  Indeed, F-1 score does not tell us all about the model's performance. Sometimes training for more epochs could lead to lower loss values while other metrics (such as precision, recall, or F-1) no longer improve. That would still be desirable because it means the model is now more ``confident,'' in a sense, about its predictions. However, that is not the case here. As shown in Figure~\ref{fig:spacy_epoch},
after around 40 epochs the loss begins to oscillate instead of continuing downwards, suggesting that in this case training for 100 epochs does not result in a better model than only training for 20 epochs.

\begin{figure}[bht]
    \centering
    \begin{subfigure}{.43\linewidth}
      \centering
      \includegraphics[width=\linewidth]{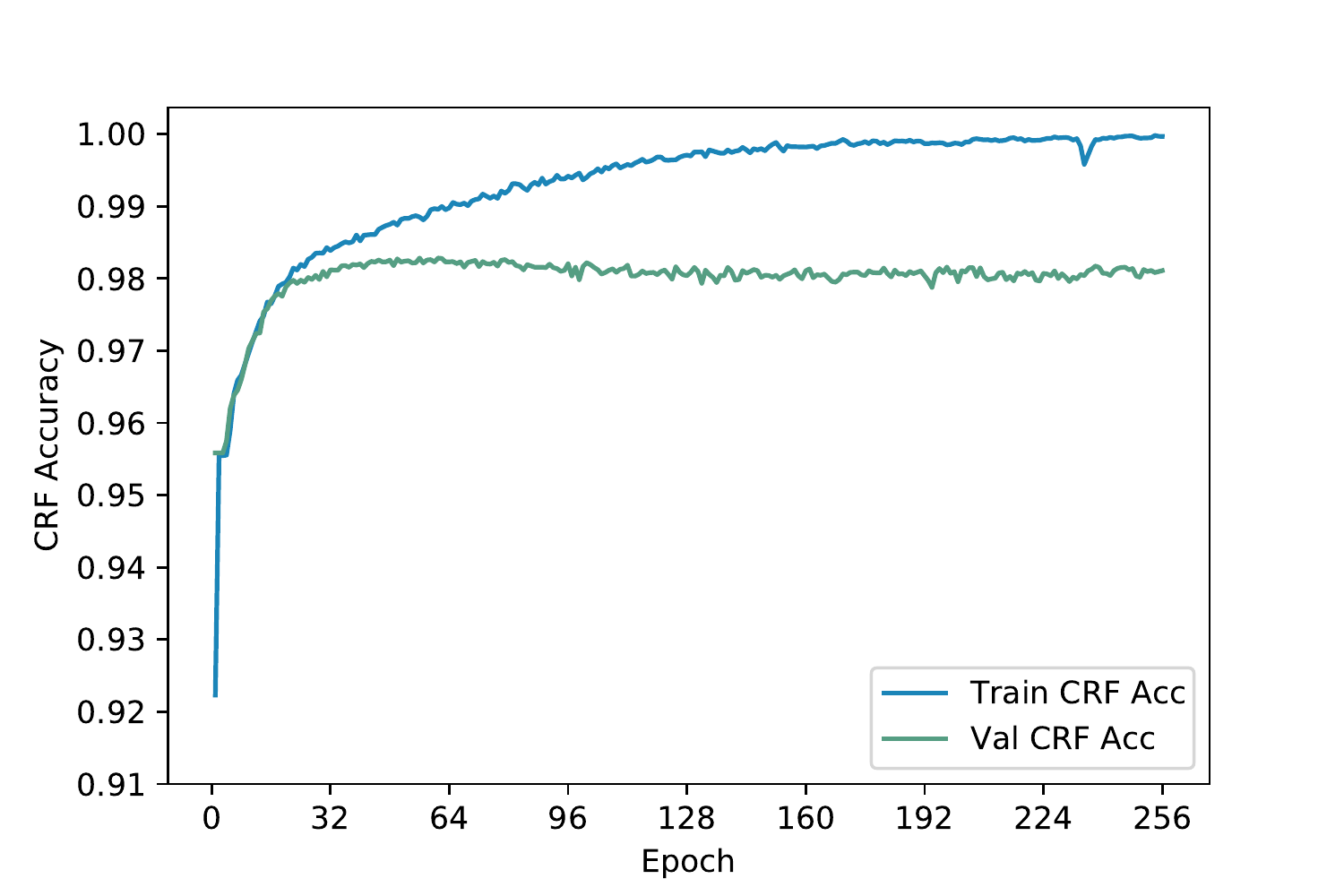}
      \caption{Accuracy}
      \label{fig:lstm_epoch_accu}
    \end{subfigure}%
    ~
    \begin{subfigure}{.43\linewidth}
      \centering
      \includegraphics[width=\linewidth]{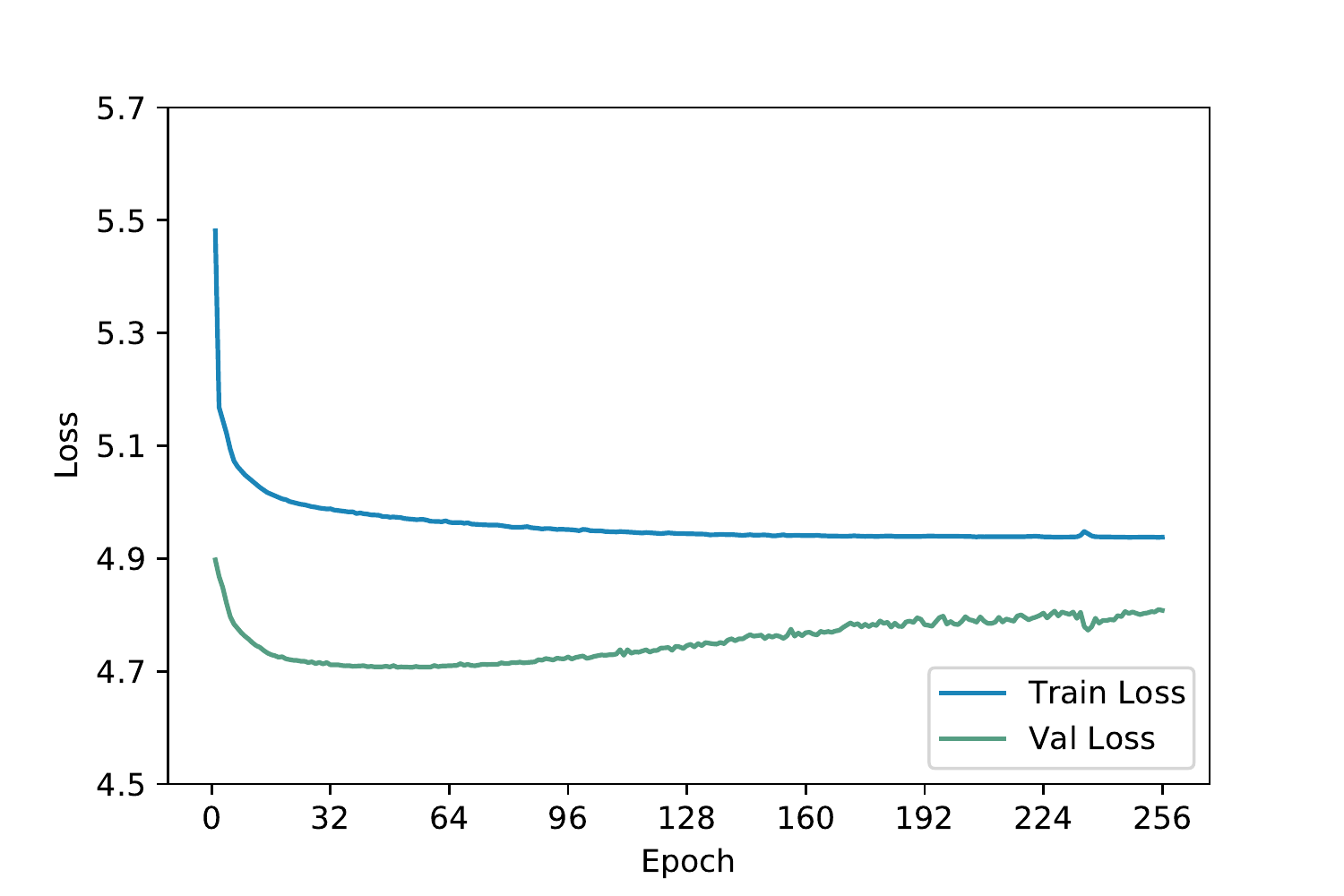}
      \caption{Loss}
      \label{fig:lstm_epoch_loss}
    \end{subfigure}
    \caption{Training and validation accuracy and loss for Keras-LSTM model over 256 epochs.}
    \label{fig:lstm_epoch}
\end{figure}

Figure~\ref{fig:lstm_epoch} shows the progression of accuracy and loss value for the Keras-LSTM model with the initial 278 examples. In Figure~\ref{fig:lstm_epoch}(a), we see that validation  accuracy improves as training accuracy increases during the first 50 epochs. After around epoch 50, the training and validation accuracy curves diverge: the training accuracy continues to increase but the validation accuracy plateaus. This trend is suggestive of overfitting, which is corroborated by Figure~\ref{fig:lstm_epoch}(b). After about 50 epochs, the validation loss curve turns upwards. Hence we choose to limit the training epochs to 64. After each epoch, if a lower validation loss is achieved, the current model state is saved. After 64 epochs, we test the model with the lowest validation loss on the withheld test set.

\section{Evaluating Model and Human Performance}
\label{sec:ner_models}

We conducted experiments to compare the performance of the SpaCy and Keras-LSTM NER models; compare the performance of the models against humans; determine how training data influences model performance; and analyze human and model errors. 

\subsection{Performance of SpaCy and Keras NER Models}
\label{sec:spacy_keras_performance}
We used the collected data of Section~\ref{sec:data_collection}
to train both the SpaCy and Keras-LSTM NER models of Section~\ref{sec:models} to recognize and extract drug-like molecules in text. 
We find that the trained \texttt{en\_core\_web\_md} SpaCy model achieved a best F-1 score of 77.3\%,
while the trained Keras-LSTM model achieved a best F-1 score of 80.5\%,
somewhat outperforming SpaCy.

As shown in Figure~\ref{fig:num_of_ex}, the SpaCy model performs better than the Keras-LSTM model when trained with small amounts of training data---perhaps because of the different mechanisms employed by the two methods to generate numerical representations for words. SpaCy's built-in language model, pre-trained on a general corpus of blog posts, news, comments, etc., gives it some knowledge about commonly used words in English, which are likely also to appear in a scientific corpus. On the other hand, the Keras-LSTM model uses custom word embeddings trained solely on an input corpus, which provides it with better understanding of multi-sense words, especially those that have quite different meanings in a scientific corpus. However, without enough raw data to draw contextual information from, custom word embeddings can not accurately reflect the meaning of words. 

\subsection{Comparison Against Human Performance}
Recognizing drug-like molecules is a difficult task even for humans, especially non-medical professionals (such as our non-expert annotators). To assess the accuracy of the annotators, we asked three people to examine 96 paragraphs, with their associated labels, selected at random
from the labeled examples. Two of these reviewers had been involved in creating the labeled dataset; the third had not.
For each paragraph, each reviewer decided independently whether each 
drug molecule entity was labeled correctly (a true positive),
was labeled as a drug when it was not (a false positive),
or was not labeled (a false negative).
If all three reviewers agreed in their opinions on a paragraph
(the case for 88 of the 96 paragraphs), we accepted their opinions;
if they disagreed (the case for eight paragraphs), 
we engaged an expert.

This process revealed a total of 257 drug molecule entities
in the 96 paragraphs, of which the annotators labeled 201 correctly (true positives), labeled 49 incorrectly (false positives), and missed 34 (false negatives). The numbers of true positives and false negatives do not sum up to the total number of drug molecules because in some cases an annotator labeled not to a drug entity but the entity plus extra preceding or succeeding word or punctuation mark (e.g. ``sofosbuvir,'' instead of ``sofosbuvir'') and we count such occurrences as false positives rather than false negatives. In this evaluation, the non-expert annotators achieved an F-1 score of 82.9\%, which is comparable to the 80.5\% achieved by our automated models, as shown in Figure~\ref{fig:num_of_ex}.
In other words, our models have performance on par with that of non-expert humans.

\subsection{Effects of Training Data Quality on Model Performance}
\label{sec:diff_data_source}

We described in the previous section how review of 96 paragraphs labeled by the non-expert annotators revealed an error rate of about 20\%. This raises the question of whether model performance could be improved with better training data. 
To examine this question, we compare the performance of our models when trained on original vs.\ corrected data. 
As we only have 96 corrected paragraphs, 
we restrict our training sets to those 96 paragraphs in each case.

We sorted the 96 paragraphs in both datasets so that they are considered in the same order. Then, we split each dataset into five subsets for K-fold cross validation (\emph{K}=5), with the first four subsets having 19 paragraphs each and the last subset having 20. Since \emph{K} is set to five, the SpaCy and Keras models are trained five times. In the \emph{i}-th round, each model is trained on four subsets (excluding the \emph{i}-th) of each dataset. The \emph{i}-th subset of the corrected dataset is used as the test set. The \emph{i}-th subset of the original dataset is not used in the \emph{i}-th round.

We present the K-fold cross validation results in Tables~\ref{tab:spacy_kfold} and \ref{tab:keras_kfold}. The models performed reasonably well when trained on the original dataset, with an average F-1 score only 2\% less than that achieved with the corrected labels. Given that the expert input required for validation is hard to come by, we believe that using non-expert reviewers is an acceptable tradeoff and probably the only practical way to gather large amounts of training data.

\begin{table}[thb]
    \caption{K-fold (K=5) validation of the SpaCy model on 96 paragraphs with original vs.\ corrected labels. The first five rows in each table are the results of each fold; the last row is the average F-1 score of the five folds.}
    \label{tab:spacy_kfold}
    \hspace{0.15\textwidth}
    \centering
    \begin{subtable}[t]{0.3\textwidth}
        \caption{Original labels}
        \label{tab:spacy_kfold_reviewer}
        \centering
        \begin{tabular}{|c|c|c|}
            \hline
            \rowcolor{Gray}
            Precision & Recall & F-1  \\ \hline
            100.0     & 25.6   & 40.7 \\ \hline
            93.8      & 50.6   & 65.7 \\ \hline
            93.0      & 63.5   & 75.5 \\ \hline
            76.5      & 45.2   & 57.1 \\ \hline
            98.9      & 92.3   & 95.5 \\ \hline
            Average   &        & 66.9 \\ \hline
        \end{tabular}
    \end{subtable}
    \hspace{\fill}
    \begin{subtable}[t]{0.3\textwidth}
        \caption{Corrected labels}
        \label{tab:spacy_kfold_expert}
        \centering
        \begin{tabular}{|c|c|c|}
            \hline
                        \rowcolor{Gray}
            Precision & Recall & F-1  \\ \hline
            100.0     & 20.9   & 34.6 \\ \hline
            88.9      & 53.9   & 67.1 \\ \hline
            92.0      & 73.0   & 81.4 \\ \hline
            68.6      & 61.4   & 64.8 \\ \hline
            99.2      & 92.1   & 95.5 \\ \hline
            Average   &        & 68.7 \\ \hline
        \end{tabular}
    \end{subtable}
    \hspace{0.15\textwidth}
\end{table}

\begin{table}[thb]
    \caption{K-Fold (K=5) validation of the Keras-LSTM model on 96 paragraphs with original vs.\  corrected labels. The first five rows in each table are the results of each fold; the last row is the average F-1 score of the five folds.}
    \label{tab:keras_kfold}
    \hspace{0.15\textwidth}
    \centering
    \begin{subtable}[t]{0.3\textwidth}
        \caption{Original labels}
        \label{tab:keras_kfold_reviewer}
        \centering
        \begin{tabular}{|c|c|c|}
            \hline
                        \rowcolor{Gray}
            Precision & Recall & F-1  \\ \hline
            88.5      & 53.5   & 66.7 \\ \hline
            70.6      & 57.8   & 63.6 \\ \hline
            70.0      & 67.7   & 68.9 \\ \hline
            80.8      & 55.3   & 65.6 \\ \hline
            78.1      & 56.8   & 65.8 \\ \hline
            Average   &        & 66.1 \\ \hline
        \end{tabular}
    \end{subtable}
    \hspace{\fill}
    \begin{subtable}[t]{0.3\textwidth}
        \caption{Corrected labels}
        \label{tab:keras_kfold_expert}
        \centering
        \begin{tabular}{|c|c|c|}
            \hline
                        \rowcolor{Gray}
            Precision & Recall & F-1  \\ \hline
            80.6      & 69.1   & 74.4 \\ \hline
            90.2      & 59.1   & 71.4 \\ \hline
            82.9      & 47.5   & 60.4 \\ \hline
            80.8      & 51.2   & 62.7 \\ \hline
            75.0      & 67.9   & 71.3 \\ \hline
            Average   &        & 68.0 \\ \hline
        \end{tabular}
    \end{subtable}
    \hspace{0.15\textwidth}
\end{table}

\subsection{Analysis of Human and Model Errors}
Finally, we explore the contexts in which human reviewers and models make mistakes. Specifically, we study the tokens that appear most frequently near to incorrectly labeled entities. To investigate the effects of immediate and long-distance context, we control, as \emph{window size}, the maximum distance between a token and a entity for that token to be considered as ``context'' for that entity.

One difficulty with this analysis is that the most frequent tokens identified in this way were mostly stop words or punctuation marks. For instance, when the window size is set to three, the 10 most frequent tokens around mislabeled words are, in descending order, ``comma(,),'' ``and,'' ``mg,'' ``period(.),'' ``right parenthesis()),'' ``with,'' ``of,'' ``left parenthesis((),'' ``is,'' and ``or.'' Only ``mg'' is neither a stop word nor punctuation mark.

Those tokens provide little insight as to why human reviewers might have made mistakes, and furthermore are unlikely to have influenced reviewer decisions. 
Thus we exclude stopwords and punctuation marks when providing,
in Table~\ref{tab:human_incorrect_freq_tokens_no_stop}, lists of the 10 most frequent tokens
within varying window sizes of words that were  \emph{incorrectly} identified as molecules by human reviewers.

We see that there are indeed several deceptive contextual words. With a window size of one, the 10 most frequent tokens include ``oral,'' ``dose,'' and ``intravenous.'' It is understandable that an untrained reviewer might label as drugs words that immediately precede or follow such context words.
Similar patterns can be seen for window sizes of three and five. 
Without background knowledge to draw from, non-experts are more likely to rely on their experience gained from labeling previous paragraphs. One may hypothesize that after the reviewers have seen a few dozen to a few hundred paragraphs, those deceptive contextual words must have left a deep impression, so that when those words re-appear they are likely to label the strange unknown word close to them as a drug.

\begin{table}[h]
    \caption{The 10 most frequent tokens, excluding stopwords and punctuation marks, within various window sizes around entities incorrectly labeled by human reviewers.}
    \label{tab:human_incorrect_freq_tokens_no_stop}
\centering
    \begin{tabular}{|c|l|c|l|c|l|c|}
        \hline
                    \rowcolor{Gray}
        & \multicolumn{2}{c|}{window size = 1} & \multicolumn{2}{c|}{window size = 3} & \multicolumn{2}{c|}{window size = 5} \\ \hline
                    \rowcolor{Gray}
        \# & \multicolumn{1}{c|}{Token}  & Count  & \multicolumn{1}{c|}{Token}  & Count  & \multicolumn{1}{c|}{Token}  & Count  \\ \hline\hline
        1  & 300                         & 4      & mg                          & 19     & mg                          & 23     \\ \hline
        2  & oral                        & 3      & once/day                    & 7      & daily                       & 8      \\ \hline
        3  & dose                        & 3      & treatment                   & 6      & treatment                   & 8      \\ \hline
        4  & intravenous                 & 2      & 300                         & 5      & once/day                    & 7      \\ \hline
        5  & 500                         & 2      & treated                     & 4      & 300                         & 7      \\ \hline
        6  & intravenously               & 1      & oral                        & 4      & oral                        & 6      \\ \hline
        7  & include                     & 1      & once                        & 4      & recipients                  & 5      \\ \hline
        8  & Both                        & 1      & dose                        & 4      & treated                     & 4      \\ \hline
        9  & resistance                  & 1      & cidofovir                   & 3      & twice                       & 4      \\ \hline
        10 & treatment                   & 1      & resistance                  & 3      & include                     & 4      \\ \hline
    \end{tabular}
\end{table}

To investigate this hypothesis, we also explored the most frequent words around drug entities that are \emph{correctly} labeled by human reviewers:
see Table~\ref{tab:human_correct_freq_tokens}. Interestingly, we found overlaps between the lists in Tables~\ref{tab:human_incorrect_freq_tokens_no_stop} and \ref{tab:human_correct_freq_tokens}: in all, three, four, and two overlaps for window sizes
of one, three, and five, respectively, when treating all numerical values as identical.
This finding supports our hypothesis that those frequent words around real drug entities may confuse human reviewers when they appear around non-drug entities.

\begin{table}[h]
    \caption{The 10 most frequent tokens, excluding stop words and punctuation marks, within various window sizes around entities correctly labeled by human reviewers.}
    \label{tab:human_correct_freq_tokens}
\centering
\begin{tabular}{|c|l|c|l|c|l|c|}
    \hline
                \rowcolor{Gray}
       & \multicolumn{2}{c|}{window size = 1} & \multicolumn{2}{c|}{window size = 3} & \multicolumn{2}{c|}{window size = 5} \\ \hline
                   \rowcolor{Gray}
    \# & \multicolumn{1}{c|}{Token}  & Count  & \multicolumn{1}{c|}{Token}  & Count  & \multicolumn{1}{c|}{Token}  & Count  \\ \hline\hline
    1  & resistance                  & 176    & Tetracycline                & 230    & Tetracycline                & 230    \\ \hline
    2  & treatment                   & 9      & resistance                  & 177    & resistance                  & 178    \\ \hline
    3  & mM                          & 4      & Trimethoprim                & 118    & Trimethoprim                & 118    \\ \hline
    4  & oral                        & 3      & treatment                   & 11     & treatment                   & 14     \\ \hline
    5  & after                       & 3      & 20$\sim$                    & 7      & 20$\sim$                    & 8      \\ \hline
    6  & analogue                    & 3      & Figure                      & 5      & placebo                     & 7      \\ \hline
    7  & responses                   & 3      & concentration               & 5      & effects                     & 6      \\ \hline
    8  & antibiotics                 & 2      & compared                    & 4      & Figure                      & 6      \\ \hline
    9  & exposure                    & 2      & 100                         & 4      & KLK5                        & 6      \\ \hline
    10 & pharmacokinetics            & 2      & mM                          & 4      & matriptase                  & 6      \\ \hline
    \end{tabular}
\end{table}

We repeat this comparison of context words around human and model errors while considering stopwords and punctuation marks.
Tables~\ref{tab:human_incorrect_freq_tokens} and~\ref{tab:keras_incorrect_freq_tokens}
show the 20 most frequent tokens in each case.
We see that 20--25\% of the tokens in Table~\ref{tab:human_incorrect_freq_tokens},
but only 5--10\% of those in Table~\ref{tab:keras_incorrect_freq_tokens},
are \emph{not} stop words or punctuation marks. As the model only learns its word embeddings from the input text, if a token often co-occurs with drug entities in the training corpus the model will treat it as an indication of drug entities near its presence, regardless of whether or not it is a stopword.
This apparently leads the model to make incorrect inferences.
Humans, on the other hand, are unlikely to think that stopword such as ``the'' is indicative of drug entities, no matter how frequently they appear together. 

\begin{table}[h]
    \caption{The 20 most frequent tokens, including stop words and punctuation marks, within various window sizes around entities incorrectly labeled by human reviewers. Words that are neither stop words nor punctuation words are in boldface.}
    \label{tab:human_incorrect_freq_tokens}
\centering
    \begin{tabular}{|c|l|c|l|c|l|c|}
    \hline
                \rowcolor{Gray}
       & \multicolumn{2}{c|}{window size = 1} & \multicolumn{2}{c|}{window size = 3} & \multicolumn{2}{c|}{window size = 5} \\ \hline
                   \rowcolor{Gray}
    \# & \multicolumn{1}{c|}{Token}  & Count  & \multicolumn{1}{c|}{Token}  & Count  & \multicolumn{1}{c|}{Token}  & Count  \\ \hline\hline
    1  & ,                           & 27     & ,                           & 49     & ,                           & 74     \\ \hline
    2  & .                           & 14     & and                         & 21     & .                           & 28     \\ \hline
    3  & and                         & 14     & \textbf{mg}                 & 19     & and                         & 28     \\ \hline
    4  & with                        & 8      & .                           & 18     & \textbf{mg}                 & 23     \\ \hline
    5  & (                           & 7      & )                           & 13     & )                           & 21     \\ \hline
    6  & is                          & 7      & with                        & 10     & of                          & 18     \\ \hline
    7  & of                          & 6      & of                          & 10     & (                           & 17     \\ \hline
    8  & was                         & 6      & (                           & 9      & with                        & 13     \\ \hline
    9  & or                          & 4      & is                          & 9      & to                          & 12     \\ \hline
    10 & \textbf{300}                & 4      & or                          & 7      & the                         & 11     \\ \hline
    11 & \textbf{oral}               & 3      & \textbf{once/day}           & 7      & a                           & 11     \\ \hline
    12 & has                         & 3      & a                           & 7      & in                          & 10     \\ \hline
    13 & to                          & 3      & the                         & 6      & is                          & 9      \\ \hline
    14 & {[}                         & 3      & was                         & 6      & or                          & 9      \\ \hline
    15 & \textbf{dose}               & 3      & to                          & 6      & \textbf{daily}              & 8      \\ \hline
    16 & \textbf{intravenous}        & 2      & \textbf{treatment}          & 6      & was                         & 8      \\ \hline
    17 & in                          & 2      & in                          & 5      & \textbf{treatment}          & 8      \\ \hline
    18 & may                         & 2      & \textbf{300}                & 5      & {]}                         & 8      \\ \hline
    19 & \textbf{500}                & 2      & be                          & 4      & \textbf{once/day}           & 8      \\ \hline
    20 & a                           & 2      & \textbf{treated}            & 4      & were                        & 7      \\ \hline
    \end{tabular}
\end{table}

\begin{table}[thb]
    \caption{The 20 most frequent tokens, including stop words and punctuation marks, within various window sizes around entities incorrectly labeled by the Keras model. Words that are neither stop words nor punctuation words are in boldface.}
    \label{tab:keras_incorrect_freq_tokens}
\centering
    \begin{tabular}{|c|l|c|l|c|l|c|}
    \hline
                \rowcolor{Gray}
       & \multicolumn{2}{c|}{window size = 1} & \multicolumn{2}{c|}{window size = 3} & \multicolumn{2}{c|}{window size = 5} \\ \hline
                   \rowcolor{Gray}
    \# & \multicolumn{1}{c|}{Token}  & Count  & \multicolumn{1}{c|}{Token}  & Count  & \multicolumn{1}{c|}{Token}  & Count  \\ \hline\hline
    1  & ,                           & 166    & ,                           & 347    & ,                           & 468    \\ \hline
    2  & (                           & 86     & and                         & 126    & and                         & 176    \\ \hline
    3  & and                         & 81     & (                           & 117    & (                           & 162    \\ \hline
    4  & of                          & 58     & )                           & 89     & )                           & 143    \\ \hline
    5  & )                           & 30     & of                          & 85     & of                          & 130    \\ \hline
    6  & to                          & 28     & the                         & 73     & the                         & 121    \\ \hline
    7  & or                          & 24     & to                          & 60     & to                          & 85     \\ \hline
    8  & .                           & 22     & .                           & 48     & in                          & 75     \\ \hline
    9  & \textbf{mM}                 & 18     & with                        & 44     & with                        & 72     \\ \hline
    10 & with                        & 17     & a                           & 41     & .                           & 68     \\ \hline
    11 & in                          & 15     & in                          & 37     & a                           & 62     \\ \hline
    12 & for                         & 15     & or                          & 35     & was                         & 52     \\ \hline
    13 & is                          & 14     & was                         & 33     & or                          & 47     \\ \hline
    14 & as                          & 14     & \textbf{mM}                 & 32     & is                          & 43     \\ \hline
    15 & the                         & 13     & is                          & 31     & for                         & 42     \\ \hline
    16 & was                         & 12     & as                          & 29     & that                        & 37     \\ \hline
    17 & that                        & 11     & that                        & 25     & \textbf{mM}                 & 36     \\ \hline
    18 & {[}                         & 11     & by                          & 22     & by                          & 35     \\ \hline
    19 & \textbf{treatment}          & 9      & for                         & 22     & as                          & 32     \\ \hline
    20 & a                           & 9      & {[}                         & 22     & were                        & 31     \\ \hline
    \end{tabular}
\end{table}

\section{Applying the Trained Models}\label{sec:apply}

After training the models with the labeled examples, we applied the trained models to the entire CORD-19 corpus (2020-10-04 version with \num{198875} articles) to identify potential drug-like molecules. 
Processing a single article takes only a few seconds;
we adapted our models to use data parallelism to enable rapid processing of these many articles.

\begin{figure}[b!]
    \centering
    \begin{subfigure}{.43\linewidth}
      \centering
      \includegraphics[width=\linewidth,trim=3mm 2mm 15mm 11mm,clip]{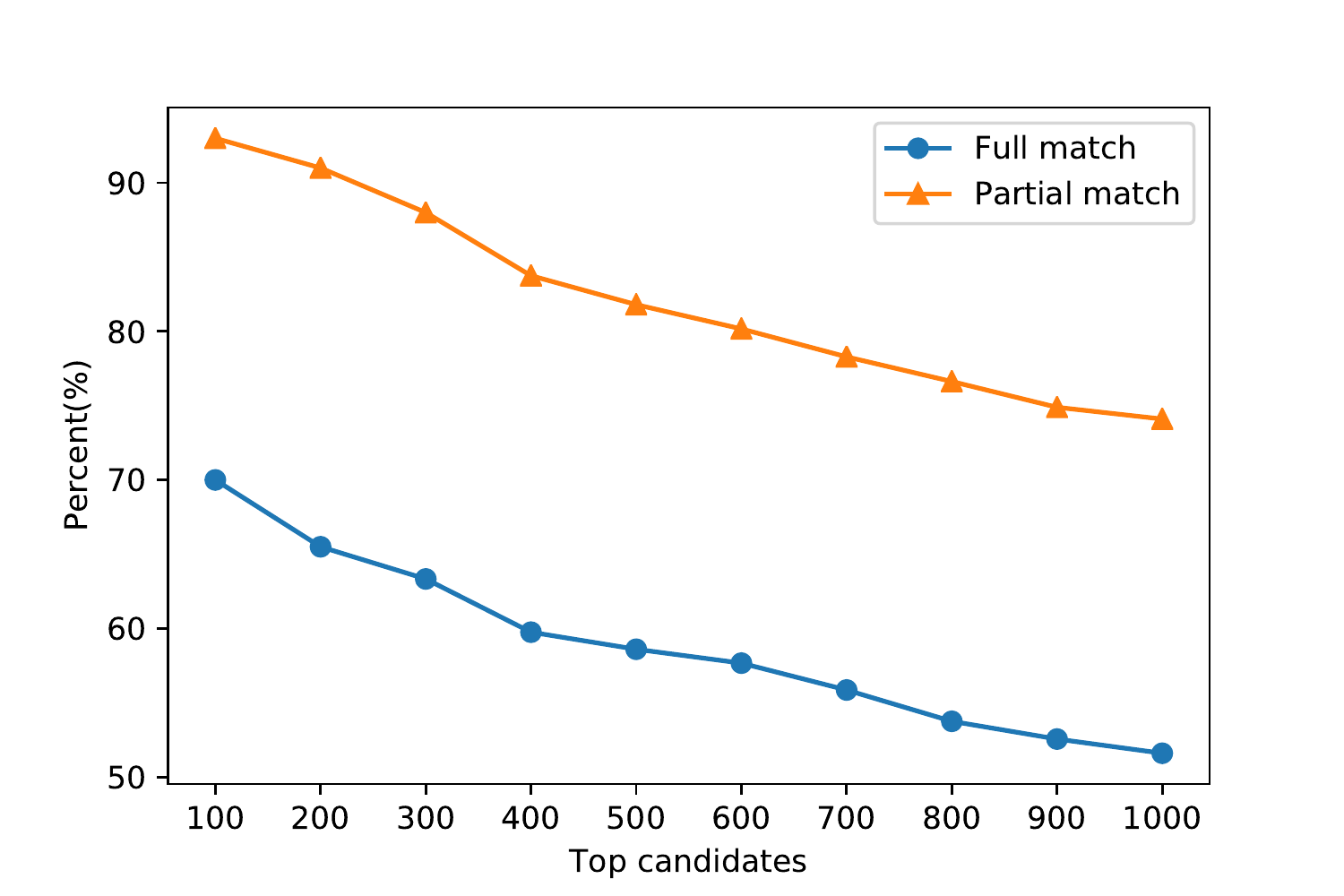}
      \caption{All Entities}
      \label{fig:cmp_to_drugbank_all_entities}
    \end{subfigure}%
    ~
    \begin{subfigure}{.43\linewidth}
      \centering
      \includegraphics[width=\linewidth,trim=3mm 2mm 15mm 11mm,clip]{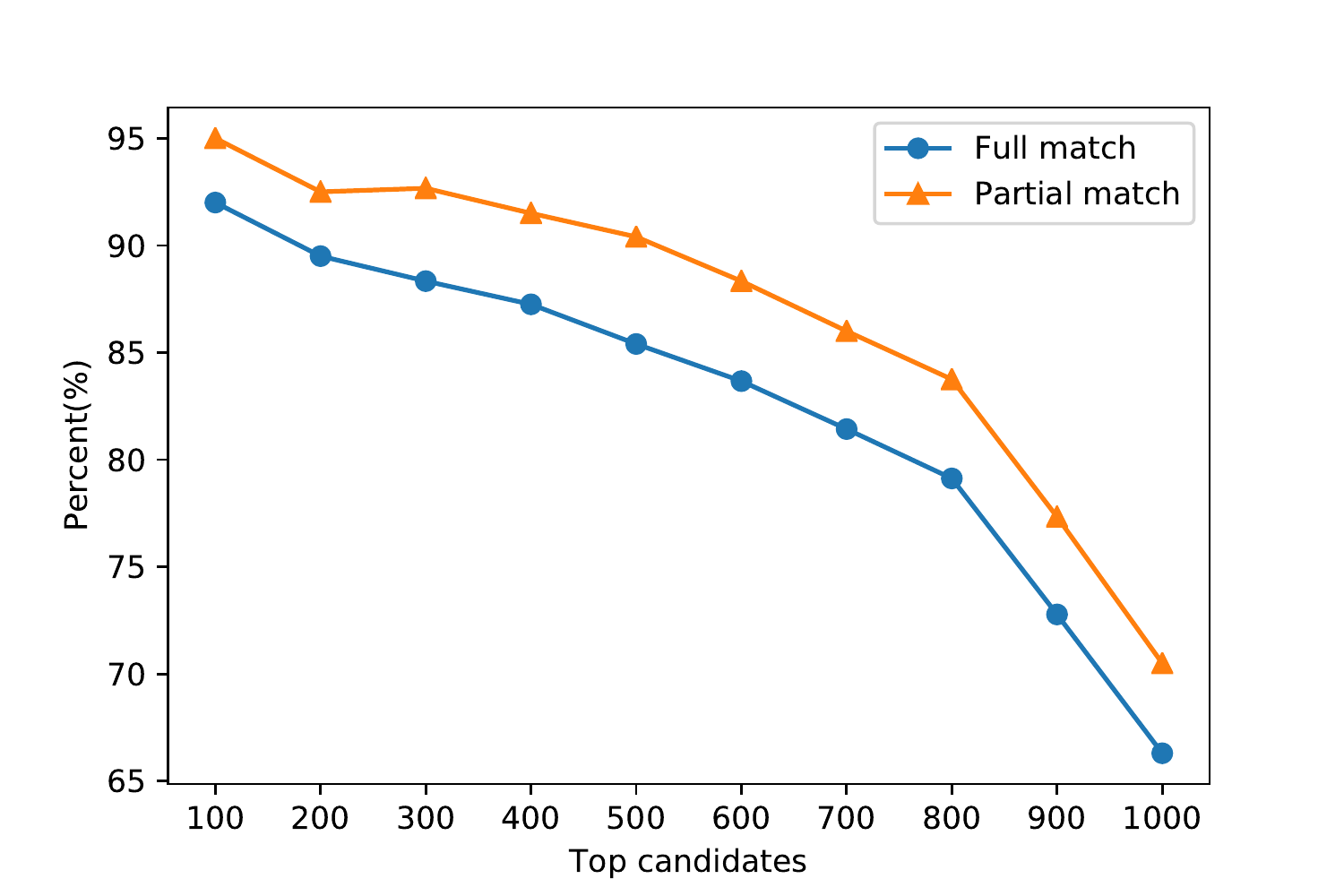}
      \caption{Balanced Entities}
      \label{fig:cmp_to_drugbank_balanced_entities}
    \end{subfigure}
    \caption{Percentage of detected entities that are also found in DrugBank}
    \label{fig:cmp_to_drugbank}
\end{figure}

We ran the SpaCy model on two Intel Skylake 6148 processors with a total of 40 CPU cores; this run took around 80 core-hours and extracted \num{38472} entities. We ran the Keras model on four NVidia Tesla V100 GPUs; this run took around 40 GPU-hours and extracted \num{121680} entities. 
We recorded for each entity the number of the times that it has been recognized by each model, and used those numbers as a voting mechanism to further determine which entities are the most likely to be actual drugs. In our experiments, ``balanced'' entities (i.e., those whose numbers of detection by the two models are within a factor of 10 of each other) are most likely to appear in the DrugBank list. As shown in Figure~\ref{fig:cmp_to_drugbank}, we sorted all extracted entities in descending order by their total number of detection by both models, and when comparing the the top 100 entities to DrugBank, only 77\% were exact matches to drug names or aliases, or 86\% if we included partial matches (i.e., the extracted entity is a word within a multi-word drug name or alias in DrugBank). In comparison, among the top 100 ``balanced'' entities, 88\% were exact matches to DrugBank, or 91\% with partial matches.

Although DrugBank provides a reference metric to evaluate the results, it is not an exhaustive ontology. For instance, remdesivir, a drug that has been proposed as a potential cure for COVID-19, is not in DrugBank.  We manually checked the top 50 ``balanced'' and top 50 ``imbalanced'' entities not matched to DrugBank, and found that 70\% in the ``balanced'' list are actual drugs, but only 26\% in the ``imbalanced'' list.  Looking at the false positives in these top 50 lists, the ``balanced'' false positives are often understandable. For example, in the sentence ``ELISA plate was coated with \dots and then treated for 1h at 37.8C with dithiothreitol \dots'', the model mistook the redox reagent \emph{dithiothreitol} for a drug entity, probably due to its context ``treated with.'' On the other hand, we found no such plausible explanations for the false positives in the ``imbalanced'' list, where most false positives are chemical elements (e.g., silver, sodium), amino acids (e.g., cysteine, glutamine), or proteins (e.g., lactoferrin, cystatin).

Finally, we compared our extraction results to the drugs being used in clinical trials, as listed on the U.S.\ National Library of Medicine website~(\cite{clinicaltrials}). We queried the website with ``covid'' as the keyword and manually screened the returned drugs in the ``Interventions'' column to remove stopwords (e.g., tablet, injection, capsule) and dosage information (e.g., 2.5mg, 2.5\%) and only kept the drug names. Then we compared the top 50 most frequently appeared drugs to the automatically extracted drugs from literature. The ``balanced'' entities extracted by both models matched to 64\% of the top 50 drugs in clinical trial, whereas the ``imbalanced'' entities only matched to 6\% in the same list.

The results discussed here are available in the repository described in Section~\ref{sec:data_availability}.

\section{Data Availability and Formats}
\label{sec:data_availability}

We have made our annotated training data, trained models, and the results of applying the models to the CORD-19 corpus publicly available online. \cite{lit-db}.

In order to facilitate training of various models, we published the training data in two formats---an unsegmented version in line-delimited JSON (JSONL) format, and a segmented version in Comma Separated Value (CSV) format. 
The JSONL format contains the most comprehensive information that we have collected on the paragraphs in the dataset. 
We choose JSONL format rather than a JSON list because it allows for the retrieval of objects without having to parse the entire file. 
A JSON object in the JSONL file has the following structure:
\begin{itemize}
    \item \texttt{text}: The original paragraph stored as a string without any modification.
    \item \texttt{tokens}: The list of tokens from \texttt{text} after tokenization.
    \begin{itemize}
        \item   \texttt{text}: The text of the token as a string.
        \item   \texttt{start}: The index of the first character of the token in \texttt{text}.
        \item   \texttt{end}: The index of the first character after the token in \texttt{text}.
        \item   \texttt{id}: Zero-based numbering of the token.
    \end{itemize}
    \item \texttt{spans}: The list of spans (sequences of tokens) that are labeled as named entities (drugs)
    \begin{itemize}
        \item   \texttt{start}: The index of the first character of the span in \texttt{text}.
        \item   \texttt{end}: The index of the first character after the span in \texttt{text}.
        \item   \texttt{token\_start}: The index of the first token of the span in \texttt{text}.
        \item   \texttt{token\_end}: The index of the last token of the span in \texttt{text}.
        \item   \texttt{label}: The label of the span (``drug'')        
    \end{itemize}
        
\end{itemize}

Another commonly adopted labeling scheme for NER datasets is the ``IOB'' labeling scheme, in which the original text is first tokenized and each token is assigned a label ``I,'' ``O,'' or ``B.'' The label ``B(eginning)'' means the corresponding token is the first in a named entity. A label ``I(nside)'' is given to every token in a named entity except for the first token. All other tokens gets the label ``O(utside)'' which means they are not part of any named entity. The aforementioned JSONL data are converted according to the IOB scheme and stored in Comma Separated Value (CSV) files with one training example per line. Each line consists of two columns: a first of tokens that made up of the original texts, and a second of the corresponding IOB labels for those tokens. In addition to a different labeling scheme, the samples in the CSV files are segmented, meaning that each sentence is treated as a training sample instead of an entire paragraph. This structure aligns with that used in standard NER training sets such as CoNLL03 \cite{sang2003introduction}.

The trained SpaCy and Keras models and the results of applying the models to the \num{198875} articles in the CORD-19 corpus are also available in this GitHub repo. Additionally, the pre-trained SpaCy model is provided as a cloud service via DLHub~\cite{li147dlhub, dlhub-spacy}. (The Keras model could not be hosted there due to  compatibility issues with DLHub.) This cloud service allows researchers to apply the model to any texts they provide with as few as four lines of code.

\section{Conclusion and Future Directions}

We have presented a human-machine hybrid pipeline for collecting training data for named entity recognition models. We applied this pipeline to create an automated model for identifying drug-like molecules in COVID-19-related research papers. Our pipeline facilitated efficient use of valuable human resources by presenting human labellers only with samples that were most likely to confuse our model. 
We explored various trade-offs, including model size, number of training samples, and training epochs, to find the right balance between model performance and time-to-result. In total, human reviewers working with our pipeline validated labels for 278 bootstrap samples, \num{1000} training samples, and \num{500} test samples. 
As this work was performed in conjunction with other tasks,
we cannot accurately quantify the total effort taken to collect and annotate the above training and test samples, but it was likely around 100 person-hours.

NER models trained with these data achieved a best F-1 score of 80.5\% when evaluated on our collected test set. Our models have correctly identified 64\% of the top 50 drugs that are in clinical trials for COVID-19. The models were applied to \num{198875} articles in the CORD-19 collection, from which we identified \num{10912} molecules with potential therapeutic effects against the SARS-CoV-2 coronavirus. The extracted molecule list were subsequently given to scientists at Argonne National Laboratory to be used in computational screening pipelines. The code, model, and extraction results are publicly available. In the future, we hope to further improve NER model performance by integrating our models with more advanced language models.

\section*{Data and Code Availability}
The datasets analyzed in this study can be found in the Kaggle dataset: 
\url{https://www.kaggle.com/allen-institute-for-ai/CORD-19-research-challenge}. \\
The models used in this work and the datasets generated for this study can be found on GitHub at 
\url{https://github.com/globus-labs/covid-nlp/tree/master/drug-ner}.
The models are also available on DLHub~\cite{dlhub-spacy}.

\section*{Acknowledgements}

The data generated have been prepared as part of the nCov-Group Collaboration, a group of over 200 researchers working to use computational techniques to address various challenges associated with COVID-19.

This research was supported by the DOE Office of Science through the National Virtual Biotechnology Laboratory, a consortium of DOE national laboratories focused on response to COVID-19, with funding provided by the Coronavirus CARES Act. This research used resources of the Argonne Leadership Computing Facility, a DOE Office of Science User Facility supported under Contract DE-AC02-06CH11357. 
This work was also supported by financial assistance award 70NANB19H005 from U.S.\ Department of Commerce, National Institute of Standards and Technology as part of the Center for Hierarchical Materials Design (CHiMaD).

We are grateful to our human reviewers: India S.\ Gordon, Linda Novak, Kasia Salim, Susan Sarvey, Julie Smagacz, and Monica Orozco White.

\section*{Disclaimer}

Unless otherwise indicated, this information has been authored by an employee or employees of the UChicago Argonne, LLC, operator of the Argonne National laboratory with the U.S. Department of Energy. The U.S. Government has rights to use, reproduce, and distribute this information. The public may copy and use this information without charge, provided that this Notice and any statement of authorship are reproduced on all copies.

While every effort has been made to produce valid data, by using this data, User acknowledges that neither the Government nor UChicago Argonne LLC makes any warranty, express or implied, of either the accuracy or completeness of this information or assumes any liability or responsibility for the use of this information. Additionally, this information is provided solely for research purposes and is not provided for purposes of offering medical advice. Accordingly, the U.S.\ Government and UChicago Argonne LLC are not to be liable to any user for any loss or damage, whether in contract, tort (including negligence), breach of statutory duty, or otherwise, even if foreseeable, arising under or in connection with use of or reliance on the content displayed on this site.

\printbibliography

\end{document}